\documentclass[letterpaper, 10 pt, journal, twoside]{ieeetran}

\usepackage[pdftex]{graphicx}
\graphicspath{{../pdf/}{../jpeg/}}
\DeclareGraphicsExtensions{.pdf,.jpeg,.png}

\usepackage{amsfonts}
\usepackage{graphicx}

\usepackage[cmex10]{amsmath}
\usepackage{mathabx}
\usepackage{algorithmic}
\usepackage{array}
\usepackage{mdwmath}
\usepackage{mdwtab}
\usepackage{eqparbox}
\usepackage{url}
\usepackage{svg}
\usepackage{xcolor}
\usepackage{multirow}

\usepackage{caption}
\let\origtau\tau 
\renewcommand{\tau}{\scalebox{1.44}{$\origtau$}}

\usepackage{todonotes}

\definecolor{darkgreen}{rgb}{0, .5, 0}

\title{Learning to Simulate Tree-Branch Dynamics for Manipulation}

\author{Jayadeep Jacob$^{1,2}$, Tirthankar Bandyopadhyay$^{2}$ , Jason Williams$^{2}$ , Paulo Borges$^{2}$, Fabio Ramos$^{1,3}$       
\thanks{$^{1,2}$Jayadeep Jacob is with the School of Computer Science, The University of Sydney, NSW, Australia, and with Data61, CSIRO, Pullenvale, QLD, Australia.
        {\tt\footnotesize jjac4485@sydney.edu.au}; {\tt\footnotesize jay.jacob@data61.csiro.au}}%
\thanks{$^{2}$Tirthankar Bandyopadhyay, Jason Williams, and Paulo Borges are with the Robotics and Autonomous System Group, Data61, CSIRO, Pullenvale, QLD, Australia.
        {\tt\footnotesize (tirtha.bandy,jason.williams,paulo.borges)
        @data61.csiro.au; }}

\thanks{$^{1,3}$Fabio Ramos is with the School of Computer Science, The University of Sydney, NSW, Australia, and with the NVIDIA Corporation, Seattle, USA
        {\tt\footnotesize fabio.ramos@sydney.edu.au}}%

}

\IEEEoverridecommandlockouts
\begin{document}

\maketitle

\begin{abstract}
We propose to use a simulation driven inverse inference approach to model the dynamics of tree branches under manipulation. Learning branch dynamics and gaining the ability to manipulate deformable vegetation can help with occlusion-prone tasks, such as fruit picking in dense foliage, as well as moving overhanging vines and branches for navigation in dense vegetation. The underlying deformable tree geometry is encapsulated as coarse spring abstractions executed on parallel, non-differentiable simulators. The implicit statistical model defined by the simulator, reference trajectories obtained by actively probing the ground truth, and the Bayesian formalism, together guide the spring parameter posterior density estimation. Our non-parametric inference algorithm, based on Stein Variational Gradient Descent, incorporates biologically motivated assumptions into the inference process as neural network driven learnt joint priors; moreover, it leverages the finite difference scheme for gradient approximations. Real and simulated experiments confirm that our model can predict deformation trajectories, quantify the estimation uncertainty, and it can perform better when base-lined against other inference algorithms, particularly from the Monte Carlo family. The model displays strong robustness properties in the presence of heteroscedastic sensor noise; furthermore, it can  generalise to unseen grasp locations.

\end{abstract} 

\IEEEoverridecommandlockouts

\IEEEpeerreviewmaketitle


\section{Introduction}

\IEEEPARstart{S}{imulation} driven learning approaches have been gaining popularity to address the challenging task of manipulating deformables. In this paper, we study the deformation characteristics of a tree branch under forces exerted by a robotic arm via simulation-based parameter inference. Learning the branch dynamics through simulation has significant practical ramifications, particularly for the robotics community. First, it adds the missing physics link to visual tree reconstruction works; for example \cite{lowe2022tree}\cite{quigley2021three}, to create a complete digital twin with both perceptual features and inferred dynamical properties. Second, such a digital model reduces the reality gap, often denoted as the real-to-sim gap in the robotics context, and facilitates massively parallel policy learning, inexpensive data collection, safe exploration, and most importantly, real-time control estimation. On the application front, the presence of occlusions from branches, other fruit clusters, stems and foliage pose a significant challenge to robotic fruit harvesting tasks \cite{zhou2022intelligent}, which are typically addressed through costly environment modifications to improve visibility and reachability \cite{van2020crop}. The harvest success rates of state-of-the-art methods, in the presence of partial and complete occlusions are 50-75\% and 5\% respectively \cite{zhou2022intelligent}. Along the same lines, for a mobile robot to perform autonomous navigation in natural environments, such as forests and grasslands; where plant foliage, non-compliant branches and dense vegetation pose obstacles; the conventional approach has been to avoid them altogether, quite unlike human subjects who seamlessly couple locomotion and manipulation to interact with objects, clear the path, and navigate. Having a simulated twin with inferred dynamics can overcome these barriers to improve fruit detection rate and grasp quality in harvesting tasks, and path planning with contact around obstacles for locomotion. Finally, modelling the deformation behaviour across all branch grasp locations allows the arm to effectively trade-off the low force and high reach required at the outer edge vs the high force and low reach at the inner fork, thereby, amplifying the autonomous decision making capacity of the manipulator.

\begin{figure}[t]
\vspace{1.5mm}
    \centering
    \includegraphics[width=6.8cm]{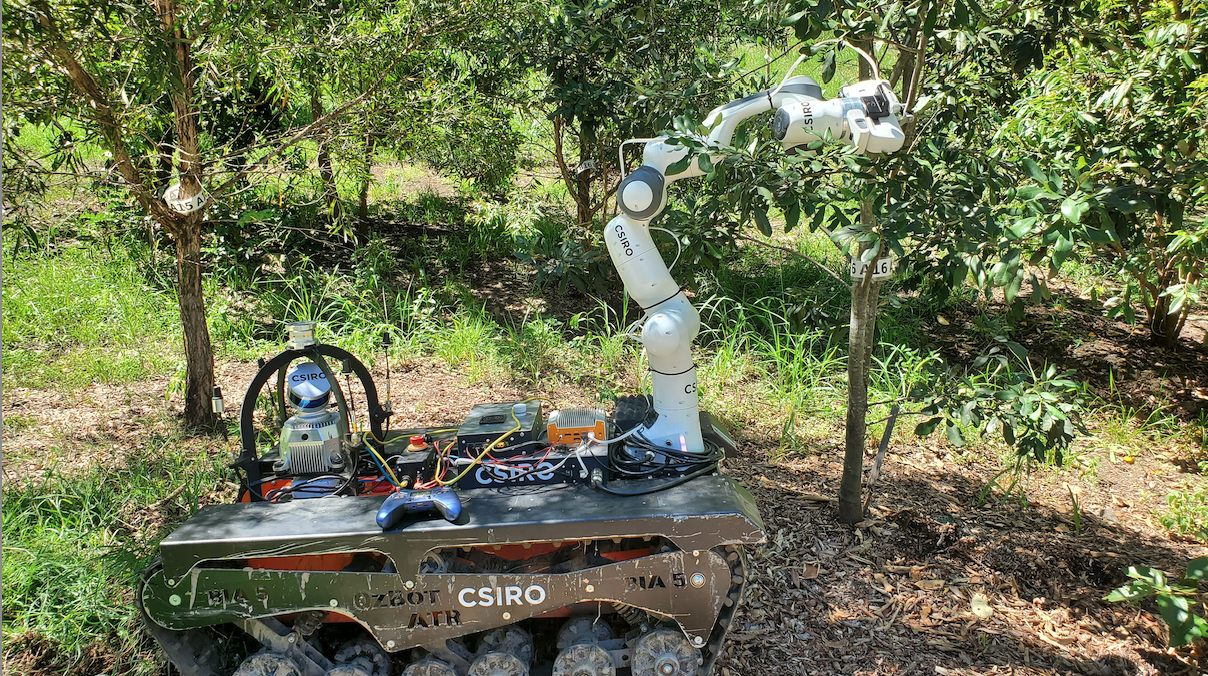}
    \caption{Autonomous navigation task: Our field robot obstructed by vegetation.}
    \label{fig:a16_franka_mobile}
\vspace*{-7mm}
\end{figure}

In general, the goal of an inverse inference approach is to estimate the latent parameters of a physical phenomenon based on its observed effects and subsequently to forecast future system behaviour. A subclass of this model, commonly known as simulation-based inference, leverages high fidelity simulations to implicitly define a statistical model and perform the inference through either frequentist or Bayesian paradigms. The parameters themselves may not necessarily have any physical meaning \cite{cranmer2020frontier}; moreover, the objective of the estimation is to predict a useful outcome rather than to accurately estimate the true parameters \cite{heiden2022probabilistic}. 

A broad outline of our workflow is as follows: to begin with, we capture the branch geometry with crude spring abstractions using purpose-built simulations executed on massively parallel, GPU based, black-box simulators. Next, we actively probe the real branches to measure the observed effects of forces. While individual twigs are rigid by nature, the additional degrees of freedom incurred due to the dynamics of coupled branches makes the problem non-linear and high dimensional. We consider the measured deformation trajectories from the real world as i.i.d samples from a ground truth distribution. Finally, guided by a Bayesian model, we use a probabilistic inference algorithm to infer the posterior density of spring parameters and to predict the branch deformation behaviour.

Overall, our work lies within the broader objective of translating the success of manipulation schemes within structured environments, guided by simulation driven inverse inference \cite{chebotar2019closing}\cite{antonova2022bayesian}, out to the natural world. However, simulation of outdoor environment is  beset by both epistemic (e.g., unaccounted factors, such as wind; model simplicity) and aleatoric (e.g., blurred vision due to limited lighting, noisy sensors) uncertainties. Therefore, optimisation models that capture point estimates of simulation parameters, from the frequentist paradigm, are not sufficient in place of an uncertainty aware Bayesian model, that we use. Although not the focus of this paper, we postulate that, on a temporal scale, each inferred posterior can feed into the next iteration as a prior, leading to faster convergence over time and that the posteriors can generalise to different branches, trees and perhaps even across species.

Our experiments demonstrate that the learnt model can estimate deformation trajectories given the end-effector forces, quantify the model uncertainty, display robustness in presence of perturbations and can perform the inference in near real time. Overall, our contributions are summarised as below:
\begin{enumerate}
\item We propose to model the dynamics of deformable branches as mass-spring abstractions on a parallel non-differentiable simulator, obtain reference trajectories by actively probing real trees, and infer the spring parameters via a probabilistic inverse inference algorithm.
\item We incorporate biologically motivated assumptions into the Bayesian inference workflow by formulating inequality constraints as differentiable and learnt joint prior distributions.
\item We demonstrate the real-to-sim transfer with two different real-world robots, operating on distinct physical trees, thereby additionally validating model independence from both arm kinematics and branch topology.
\item Finally, we show that the learnt model and the predicted deformation trajectories are robust to noise perturbations from the sensors and variations in branch grasp locations where the manipulation forces are applied.
\end{enumerate} 


\section{Related Work}
Manipulation of deformable objects, such as clothes, elastic bands, and ropes has gained much traction over the years, owing to the broad range of ensuing applications. The comprehensive surveys \cite{yin2021modeling} and \cite{arriola2020modeling}  detail established techniques to represent shapes, mange the dynamics, and apply learning strategies, all in order to  perform control and planning with deformables. While mass-spring models have been used to build coarse-grained, but computationally efficient models for suturing \cite{schulman2013case} and fabric management \cite{makris2022deformable}, its more accurate, but slower counterpart, the finite element method (FEM) models have been used for object tracking \cite{bozic2020deepdeform} and manipulating rubber tubes \cite{rambow2012autonomous}. Perhaps more in line with our own work, both mass-spring and FEM models are used in conjunction, the latter as a reference, to acquire deformation trajectories for motion planning around soft toys \cite{frank2014learning}. 

In contrast, dynamic behaviour modelling of tree branches for active manipulation is rather rare in literature. Spring based models have been used to aid visual reconstruction of trees \cite{yandun2020visual} or to animate the interaction between rain drops and branches \cite{yang2010physically}; however, the spring parameters are specified in advance. On the other hand, applying spring models to estimate branch behaviour under wind loading has been extensively studied in forestry \cite{de2008effects}; for example, to measure the overall sway of tree \cite{james2014study}. Such works depend on accurate modelling of the wind forces, can only compute parameters along the wind direction, and therefore they are unsuitable for active robotic manipulation. To the best of our knowledge, no works exist in the literature that actively manipulate tree branches to estimate the spring parameters through simulation.

On the inference front, probabilistic methods have been proposed to estimate parameters to bridge the gap between simulation and reality, the real-to-sim problem, for both rigid bodies \cite{heiden2022probabilistic} and deformables \cite{antonova2022bayesian}.  While these approaches estimate parameter posteriors through the Bayesian framework, they almost always start with an uninformative prior, a uniform prior to enforce limits, or other well known distributions to represent the subjective belief about the parameters. Weighing the inference technique alone, our approach appears closest to \cite{heiden2022probabilistic}, which introduces a constrained version of the SVGD algorithm inspired by the Lagrange multiplier optimization strategy. In comparison, we treat the simulator as a black-box, do not rely on accurate gradients for the simulation rollouts, and our inference formulation can embody inequality assumptions as prior beliefs.  


\section{Approach}

\subsection{Mass-Spring-Damper Model}

\begin{figure*} 
\begin{tabular}{cccc}
\centering
    \includegraphics[height=0.16\textwidth]{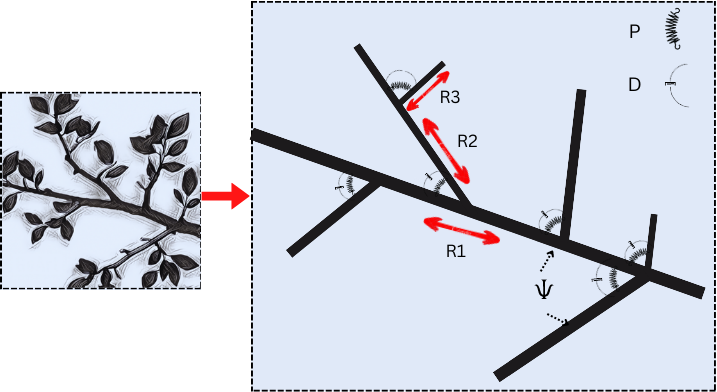}&
    \includegraphics[height=0.15\textwidth]  {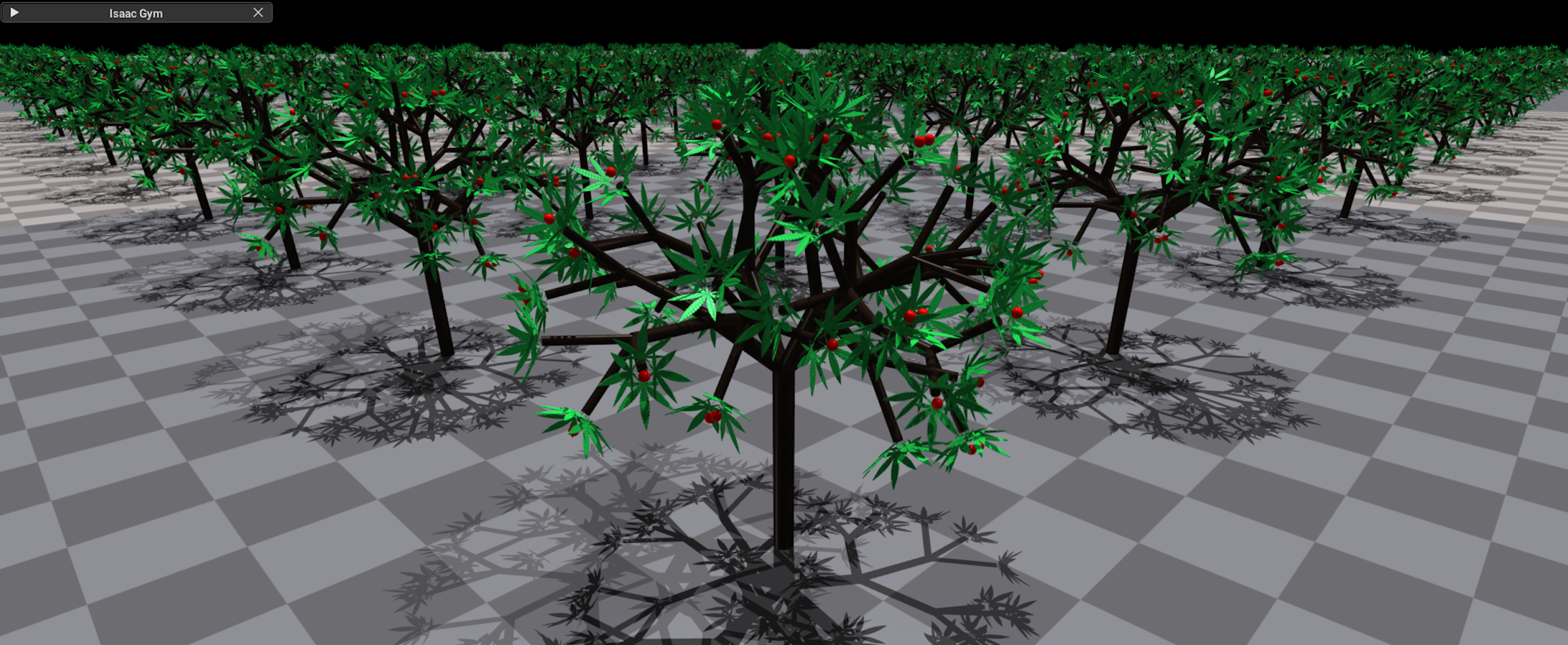} &
    \includegraphics[height=0.164\textwidth]{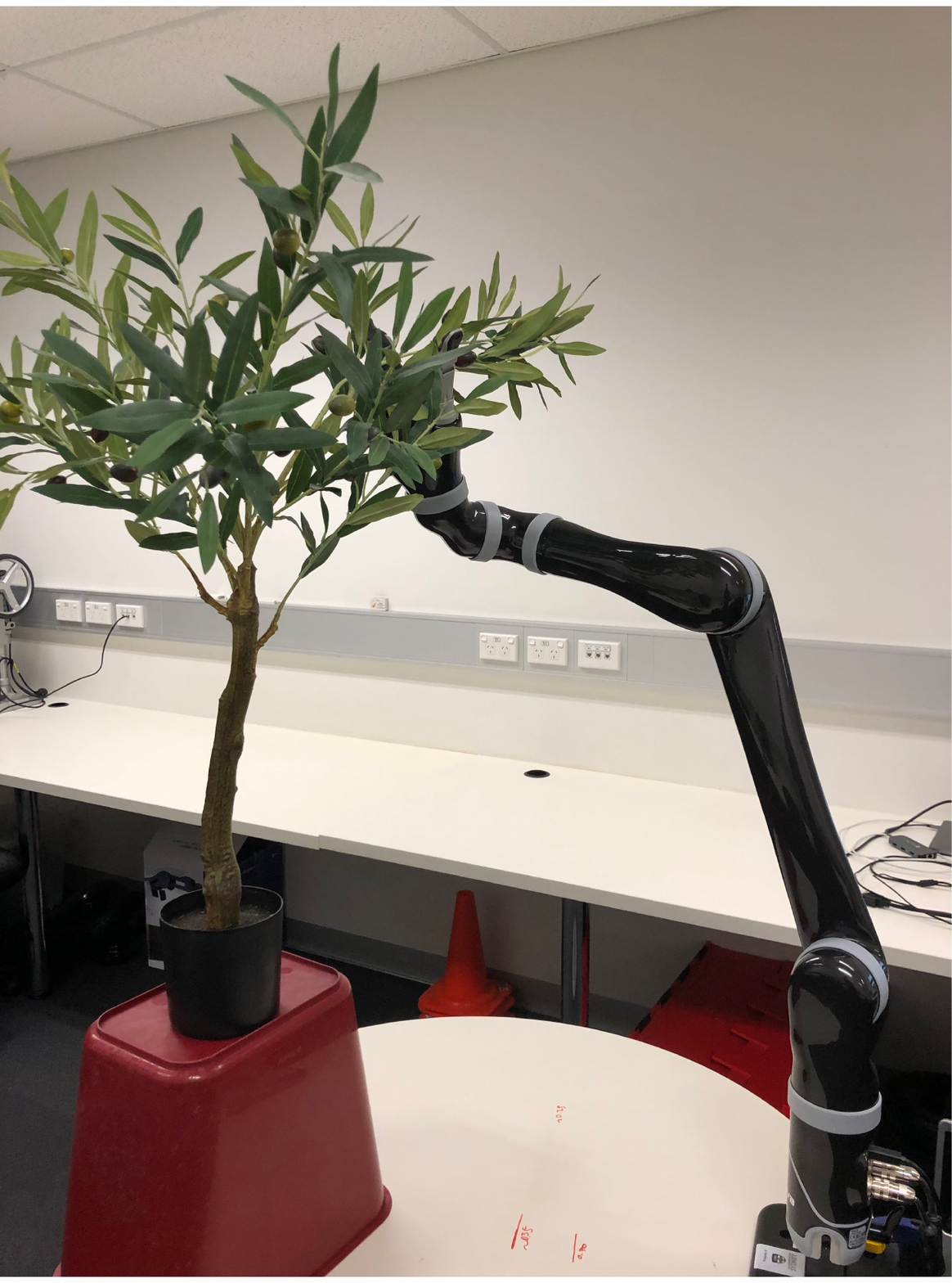}
    \includegraphics[height=0.16\textwidth]{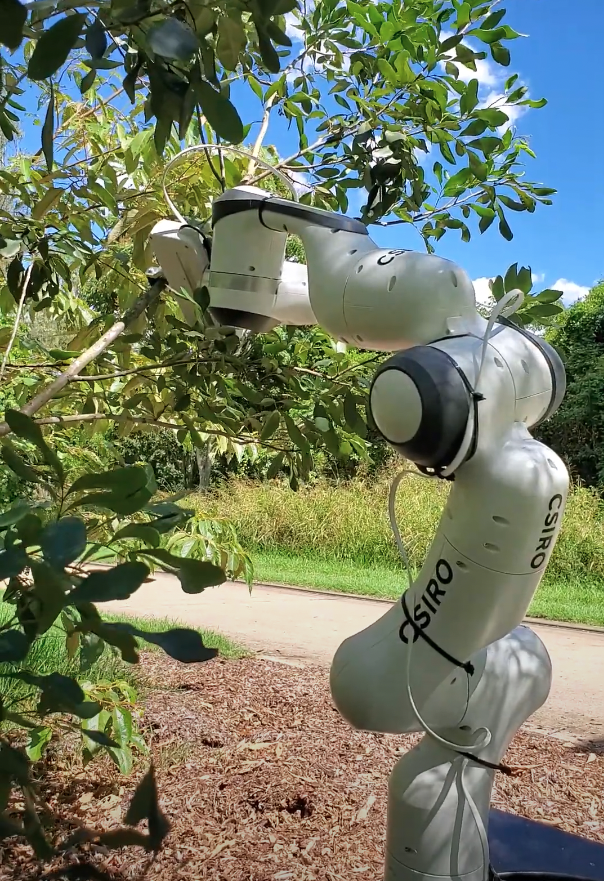} \\
     (a) & (b) & (c) 

\end{tabular}

\caption{ (a) Mass-Spring-Damper representation of a branch system (b) Parallel instances of our coarse-grain tree geometry simulations, executed on NVIDIA Isaac Gym. (c) Physical manipulator setups. Left: Kinova Jaco-2 arm on a synthetic plant.  Right: Franka-Emika Panda arm on a farm tree.  }
\label{fig:mss-sim-physical-2}
\vspace*{-5mm}
\end{figure*}

Our work utilises the well established Mass-Spring-Damper System (MSS) abstraction to simplify the intricate tree geometry. MSS models are computationally efficient and simple to implement; however, the spring constants are hard to obtain from material properties without reference models. In this work, we describe the branch structures as cylindrical links coupled with a torsional spring-damper system (Fig. \ref {fig:mss-sim-physical-2}a).

The dynamics of a single branch, and consequently that of the entire branch system, can be derived from Newton's second law of motion. Given an external torque $T_{ext}$, branch stiffness $P$, damping factor $D$, moment of inertia $J$, and  angular displacement $\psi$, this relationship can be obtained as:
\begin{equation}
T_{ext} = J \ddot{\psi} + D\dot{\psi} + P\psi
\end{equation}
We propose to model the relation $f:T_{ext} \mapsto (\psi, \dot{\psi})$, by comparing the model displacement trajectories to reference trajectories from the real world. We use physics simulators to model this MSS abstraction, which additionally helps to account for the latent attributes that affect the branch motion, such as gravity and the branch fork angle.

\subsection{Tree Simulation}  \label{tree_simulation}

Generating a digital twin for a physical entity enables cost effective data generation and robust control policy learning; moreover, GPU-accelerated simulators provide the advantage of faster training and efficient run time inference. This section describes our approach to generating crude digital tree replicas that run in parallel to perform simulator driven learning.

To begin with, we purpose-built a tool for generating tree geometry.  The geometric structures can be generated from scratch, or alternatively, it can be integrated with a perception system to create replicas of real world trees. The focus of this paper is the former, where the geometric attributes  are specified in advance. The resulting coarse tree model (Fig. \ref {fig:mss-sim-physical-2}b) takes cues from the natural world to replicate traits of real trees, such as having a fractal pattern and adhering to the cross sectional area-preservation principle \cite{minamino2014tree}, also known as the Leonardo da Vinci's rule. Furthermore, our approach is not bound by branch topology, and it can accommodate larger plant models built from Lindenmayer systems (L-system).

Subsequently, the MSS model is implemented as a multi-link structure with branches approximated as cylindrical links and branch forks represented by actuated revolute joints. The branch motion is simulated via a proportional–derivative (PD) controller governed by tunable stiffness and damping gain parameters, $\theta=\{K_{p}, K_{d}\}$. Consequently, the branch motion estimated by the PD controller is:
\begin{equation}
   T_{ctrl} =  K_{p} \psi + K_{d}\dot{\psi}, 
\end{equation}
where $T_{ctrl}$ is the controlling torque exerted by the simulated branch to offset the positional $\psi$ and velocity $\dot{\psi}$ errors. The final structure is represented with a simulator agnostic unified robotics description format (URDF). While the simplified mass-spring model is an imprecise approximation of the ground truth, particularly when compared to other more realistic finite element method (FEM) models, we show that it is sufficient to capture the multi-branch dynamics.

Finally, we choose NVIDIA Isaac Gym \cite{makoviychuk2021isaac} to simulate tree structures; specifically, its joint space PD controller to apply deformation torques and measure the resulting motion characteristics. Isaac Gym supports tensor based distributed environments, and it permits parallel execution of the inference without CPU bottlenecks. However, Isaac Gym simulations are non-differentiable; therefore, unlike \cite{heiden2022probabilistic}, throughout this paper we use finite difference methods to approximate gradients.

\subsection{Probabilistic Parameter Inference}

The key idea of our approach is to model the relationship between the force required to counteract the simulated branch resistance $T_{ctrl}$ and the resulting branch deformation trajectory, through inverse inference of  stiffness $K_{p}$ and damping $K_{d}$ gain parameters. The moment of inertia is computed by the simulator directly from the branch topology, and therefore is known a priori. The cost of deformation is established by comparing the position and velocity trajectories between the ground truth and the simulation through a loss function. The ground truth trajectories can be generated in two ways; first, as yet another independent articulated body simulation as described in section \ref{sim-to-sim}, and second using real robotic arms on physical trees, as described in section \ref{real-to-sim}. 

We define force profile $F$ as a sequence of force vectors applied on the target branch, at a location $C$, over $g$ time steps, i.e., $F:= \left \{f_{0}, f_{1}, ..., f_{g} \ | \  f \in \mathbb{R}^{3} \right \}.$ The resulting deformation at each time step $t$ is defined as the change in the position of the branch, at the force location, w.r.t its initial position, i.e., the deviation of the point $C_{t}$ from its rest position $C_{0}$. The position trajectory $\tau$ is then the sequence of such deformations, over $g$ time steps, while its derivative w.r.t time gives $\dot{\tau}$, the velocity trajectory. $\tau _{gt}$ and $\tau _{sim}$ represent the branch displacement for ground truth and simulation respectively. However, the force profile is constrained to the gravity axis (downward) as a simplification measure during experiments.

A general parameter inference problem is characterised by three terms: a parameterised simulator, a loss, and an inference framework. First, we define the simulator as a function $S$ that steps forward in time, given simulation parameters $\theta$ and force profile, to roll out  a trajectory $\tau _{sim}$ over time.
\begin{align*} 
S:(\theta, s_{i}, f_{i}) \mapsto s_{i+1} , \  \theta \in \mathbb{R}^{N} \\ 
\therefore \ \tau_{sim}^{\theta} :=  \left \{s_{0}^{\theta}, s_{1}^{\theta}, ..., s_{g}^{\theta} \right \},
\end{align*} where $N$ is the simulation parameter count. For each branch, we take $N=2$ to represent its PD coefficients, $K_{p}$ and $K_{d}$. The relationship to be modelled then is  $f:F\mapsto (\tau, \dot{\tau})$. Second, we define the loss $L$ over simulation parameters as deviations of the branch position and velocity trajectories from a ground truth.
\begin{align*} 
L(\theta) := \left \| \tau_{sim}^{\theta} - \tau_{gt} \right \|_{2} + \left \| \dot{\tau}_{sim}^{\theta} - \dot{\tau_{gt}} \right \|_{2}.
\end{align*}
Finally, similar to \cite{heiden2022probabilistic} and \cite{antonova2022bayesian}, we use Bayesian inference to quantify the uncertainty associated with the simulation parameters. The canonical Bayes' rule relates the parameter posterior probability density  $p(\theta|\tau_{sim}) $ to the corresponding likelihood $p(\tau_{sim} | \theta)$ and the prior $p_{0}(\theta)$, \\
\begin{equation} \label{simple-bayes}
p(\theta|\tau_{sim}) \  \propto \ p(\tau_{sim} | \theta) \ p_{0}(\theta). \\
\end{equation}
While point estimates of optimal parameters can be obtained as $\theta_{}^{*} =\underset{\theta \  \in \ \mathbb{R}^{N}}{\mathrm{argmin}}\,L(\theta)$, i.e., by minimising the objective function,  we instead attempt to successively sample promising regions of the posterior density $p(\theta|\tau_{sim}) $  in order to capture its multiple modalities. Furthermore, we compute the likelihood term through an exponential transformation of the loss, along the lines of simulated annealing architectures \cite{kirkpatrick1983optimization},
\begin{equation} \label{likelihood_def}
p(\tau_{sim} | \theta) := \frac{e_{}^{-L(\theta)/kT}}{Z_{\theta}} \ \ \because \ L(\theta) \geq 0,
\end{equation} where $Z_{\theta}$ is the constant set of all possible energy states, $k$ the Boltzmann constant, and $T$ the fixed annealing temperature,
\begin{equation} \label{posterior_def}
	\therefore \ p(\theta|\tau_{sim}) = \ \frac{e_{}^{-L(\theta)/kT}\ p_{0}(\theta)}{Z_{\theta} \ Z_{\origtau}},   \\
\end{equation}
 where $Z_{\origtau}$ is the normalisation constant. We do away with the intractable ${Z_{\theta}Z_{\origtau}}$ term by choosing Stein Variational Gradient Descent algorithm (SVGD) as our inference algorithm.

 While our approach can be generalized to a full 3D system, in this paper, we only consider deformations in the vertical plane, implemented by restricting the force profile to the gravity-axis, i.e., perpendicular to the ground. Deformation along other axes, such as horizontal twists can be estimated by using higher-dimensional hinges at each joint in the sequence, i.e., by increasing the $\theta$ dimensions. However, we restrict the forces to tolerable limits and leave the discontinuous dynamics of branch rupture for future work. Finally, we treat each $\tau_{sim}$ trajectory as an i.i.d sample from the underlying distribution, compute the likelihood term (\ref{likelihood_def}) over multiple trajectories, and use the log form to avoid vanishing floating points,
\begin{equation} \label{likelihood_multiple_def}
\log \ p(\tau_{sim} | \theta) := \sum_{\origtau}^{} {\frac{-L(\theta)}{kT} - \log \ Z_{\theta}}.
\end{equation}

\subsection{Stein Variational Gradient Descent} \label{svgd}

This section provides an overview of the SVGD \cite{liu2016stein} and its utility in estimating branch simulation parameters. SVGD belongs to a family of Variational Inference methods that takes an optimisation approach to simulate intractable probability distributions. SVGD has been shown to converge faster than alternatives such as Markov Chain Monte Carlo (MCMC) predominantly due to its deterministic and iterative progression, akin to the traditional Gradient Descent optimisation.

The algorithm starts with $n$ randomly initialised points in $\mathbb{R}^{N}$, termed as particles, representing draws from the target density. At each step of the iteration, the particles are pushed closer by a small step size $\epsilon \in \mathbb{R}$ as below:

\begin{equation}
\theta_{i}^{t+1} \leftarrow  \theta_{i}^{t} + \epsilon \phi_{}^{*}(\theta_{i}^{t}) \ \ \forall i=1,2,..., n,
\end{equation}
where $\theta_{i}^{t}$ represents the $i_{}^{th}$ particle at time step $t$ and $\phi_{}^{*}$ represents the optimal function that minimises the Kullback–Leibler (KL) divergence between the distribution represented by the current particle state and the expected distribution, $p(\theta|\tau_{sim}) $ in our case. However, in contrast to SGD, this minimisation is performed in a  space of functions, named the Reproducing Kernel Hilbert Space (RKHS), uniquely identified by a kernel function $k:\mathbb{R}_{}^{N} \times \mathbb{R}_{}^{N} \mapsto \mathbb{R}$. We use the popular Radial Basis Function (RBF) kernel: $k(\theta,\theta_{}^{\prime}) = \exp(-\frac{||\theta - \theta_{}^{\prime}||_{}^{2}}{2\sigma_{}^{2}})$, where the bandwidth $\sigma$ is computed using the median heuristic. Using the Stein approximation from \cite{liu2016stein} to compute the steepest direction of descent in RKHS, i.e, the optimal function $\phi_{}^{*}$: 
\[
\hat{\phi}_{}^{*}(\theta) = \frac{1}{n} \sum_{j=1}^{n}k(\theta_{j}, \theta) \nabla_{\theta_{j}}\log \ p(\theta_{j}|\tau_{sim})  + \nabla_{\theta_{j}}k(\theta_{j}, \theta).
\] 

While SVGD allows for placing the particles on a GPU and performing the operations independently, the traditional performance bottleneck has been in executing the simulations to derive the gradients. In our case, the term $\nabla_{\theta_{j}}\log \ p(\theta_{j}|\tau_{sim})$ is approximated via finite differences instead. Our simulator choice, Isaac Gym, is non-differentiable but supports efficient gradient approximations through GPU based simulations; therefore, the entire inference process runs in parallel. On substituting the prior and likelihood terms from (\ref{posterior_def}), we obtain:
\begin{multline}
\label{final_svgd}
\hat{\phi}_{}^{*}(\theta) = \frac{1}{n} \sum_{j=1}^{n} - k(\theta_{j}, \theta) \nabla_{\theta_{j}}\frac{L(\theta)}{kT} \ + \\ 
 k(\theta_{j}, \theta)\nabla_{\theta_{j}} \log \ p_{0}(\theta_{j}) + \nabla_{\theta_{j}}k(\theta_{j}, \theta).
\end{multline}
Notice that the intractable constant terms from Boltzmann distribution and from the Bayes' rule, $Z_{\theta}$ and $Z_{\origtau}$ vanish due to the gradient, which is a quintessential feature of SVGD. 

\subsection{Joint Parameter Inference}
A multi-branch joint parameter estimation problem is an extension of (\ref{simple-bayes}) to $R$ coupled branches that actively take part in the collective dynamics.
\begin{equation} \label{multi-branch-bayes}
p(\theta_{1},..., \theta_{R} |\tau_{sim}) \  \propto \ p(\tau_{sim} | \theta_{1}, ..., \theta_{R}) \ p_{0}(\theta_{1}, ..., \theta_{R}) \\
\end{equation} 
This includes the child branch which is actively probed and all its main parent branches up to the fork where any deformation is observed due to the probing. While we limit our experiments with $R=3$, the model itself has no such limitation, except for the marginal increase in computational cost per increment due to the increased dimensionality of the search space.

\subsection{Smooth Box Prior} \label{smooth_box_prior}
We use the prior $p_{0}(\theta)$ from (\ref{posterior_def}) to define a search range for the inference algorithm in order to ensure convergence within a reasonable time. Using a uniform prior results in non-differentiable regions around the prior limits. For that reason, we use the Smoothed Box prior from GPyTorch library \cite{gardner2018gpytorch} to generate a smoothed version of the uniform prior to ensure differentiability at all points. Given a box defined by the lower ($\theta_{l}$) and upper ($\theta_{u}$) bound of simulation parameters $B := \{\theta \ | \  \theta \in [\theta_{l},\  \theta_{u}],  \  \theta \in \mathbb{R}_{}^{N}\}$, and a variance $\sigma^2$, the fully differentiable smoothed box prior is approximated as:
\begin{equation*}
p_{0}(\theta) \sim \exp\big(- \frac{{}d(\theta, B)_{}^{2}} {\sqrt{2 \sigma_{}^{2}}}\big)
\end{equation*} where the distance $d$ is defined as $d(\theta, B) = \underset{{\theta' \in B}} \min \ |\theta - \theta'|$.

\subsection{Neural Network based Inequality Prior} \label{nn_ineq_prior}

Further to constraining the search range with a smooth box prior, we propose to incorporate biological assumptions on the relationship between the branch parameters into the prior $p_{0}(\theta)$ when more than one branch is involved. 

For the simplest multi-branch case, where $R=2$, i.e., one parent child pair each with a single parameter $\theta_{1}\in\mathbb{R}$ and $\theta_{2} \in  \mathbb{R}$ respectively, an inequality assumption that defines the joint parameter relationship could be:
\begin{equation} \label{simple_inequality_nn_labels}
    y(\theta_{1}, \theta_{2}) := \begin{cases} 0 & \theta_{1} \geq \theta_{2}\\-\eta & \theta_{1} < \theta_{2}\end{cases},
\end{equation}
where $\eta$ is a scaling factor and $\eta \gg 0$. We propose to train a simple one-hidden layer, feed forward, neural network regressor to model $y$, and then use the predicted $y'$ as an approximation for log of prior, i.e., $\log \ p_{0}(\theta_{1}, \theta_{2})  \approx y(\theta_{1}, \theta_{2})$.

\begin{equation} \label{simiple_biology_rule}
\therefore \  p_{0}(\theta_{1}, \theta_{2}) \approx \begin{cases} 1 & \theta_{1} \geq \theta_{2}\\e_{}^{-\eta} & \theta_{1} < \theta_{2}\end{cases},
\end{equation} implying that the child branch parameter $\theta_{2}$ ($K_{p}$ or $K_{d}$) is assumed to be lower than that of the parent $\theta_{1}$. Intuitively, from the area-preservation principle \cite{minamino2014tree}, the child branch cross-sectional area is less than that of the parent, while elasticity theory dictates that the bending stiffness is proportional to the area for a cylinder. The neural network is modelled as:
\begin{equation} \label{simple_inequality_nn}
y'(\theta_{1}, \theta_{2}) = w_{0} + \sum_{k}{w_{k}\sigma(w_{k0} + w_{k1}\theta_{1} + w_{k2}\theta_{2})},
\end{equation}
where $\sigma$ is any activation function, $k$ the number of hidden units, and $w$ the network weights. 

We justify our design choices with the following arguments: First, the regressor ensures that the predicted $y'$ is continuous and fully differentiable at all points in the domain, subject to the network activation function choice. Second, equation (\ref{simiple_biology_rule}) guarantees that $p_{0}(\theta)$ is non-negative. Third, the use of SVGD as the inference algorithm renders irrelevant the need to normalise $p_{0}(\theta)$. Therefore, $y'$ can indeed be considered as an un-normalised probability density that satisfies the differentiability requirements of the term $\nabla_{\theta_{j}} \log \ p_{0}(\theta_{j})$ from (\ref{final_svgd}). Fourth, training the model on a log scale prevents prior values from getting extremely small to machine limits, keeps it comparable to the likelihood, and therefore adequately constrains the SVGD convergence to the region of interest. Finally, using a trained model to predict joint prior opens up the possibility of learning directly from the tree geometry instead of hand coding an assumption {\em a priori}. Most importantly, we assert that this approach is consistent with the Bayesian paradigm, where assumed knowledge is expressed with priors, unlike other methods that penalise constraint violations at the likelihood -  CSVGD \cite{heiden2022probabilistic}, for example.

However, parameter estimation with (\ref{simple_inequality_nn_labels}) and (\ref{simple_inequality_nn}) generates non-trivial flat regions in the target density, thereby slowing down the inference process. A better alternative is to penalise the difference in parameters as per the prior belief. 
\begin{equation} \label{exp_biology_rule_nn_labels}
    y(\theta_{1}, \theta_{2}) := \begin{cases} 0 & \theta_{1} \geq \theta_{2}\\-\eta(\theta_{2} - \theta_{1}) & \theta_{1} < \theta_{2}\end{cases}
\end{equation}

\begin{equation} \label{exp_biology_rule}
\therefore \ p_{0}(\theta_{1}, \theta_{2}) \approx \begin{cases} 1 & \theta_{1} \geq \theta_{2}\\e_{}^{-\eta(\theta_{2} - \theta_{1})} & \theta_{1} < \theta_{2}\end{cases}
\end{equation}
where $ \eta$ is a scaling factor that determines the sharpness of the gradient and $\eta > 0 $. Note that we do not make any assumptions about the relationship between the parameters when the relationship is valid, and the penalty applies only when the constraint is violated. For the simple inequality case we implemented using (\ref{exp_biology_rule}), the reduction in search space is at  least exponential in the number of priors, indicating that the benefits of adding the prior grows exponentially for every increment in $R$.


\section{Experimental Setup}
The following subsections describe our experimental strategies and the evaluation metrics used.
\subsection{Sim-to-sim} \label{sim-to-sim}

In the most basic set up, we capture both ground truth and the replica measurements from the Isaac Gym simulator. While the geometric representation of the branch structure is same in both (as described in \ref{tree_simulation}), the execution process remains independent, i.e., no information is leaked between ground truth and inference workflows except for the force-deformation profiles. As opposed to the hardware experiments, in the sim-to-sim case, we use the simulator API to apply the deformation forces rather than using simulated arms. 

\subsection{Real-to-sim} \label{real-to-sim}

This section describes the scenario where the ground truth is captured from real world using a physical robot arm. To show that our framework is agnostic to arm kinematics, we use two independent hardware platforms (Fig. \ref {fig:mss-sim-physical-2}c) for the measurements: 1) a Kinova Jaco-2 6 DoF curved wrist manipulator, and 2) a Franka-Emika Panda 7 DoF manipulator. In the first setup, we use the Kinova manipulator to probe and deform an artificial olive tree. In contrast, we operate the mobile base mounted Franka arm on a larger, real farm tree.

A few general principles applicable to both executions are examined below. The end-effector grasp position is taken as synonymous to the branch force location $C$; therefore, the Cartesian trajectory of the end-effector acts as a proxy for the branch deformation trajectory $\tau$, at the grasp location. We command the arm using position control to deform the branch downward, in the gravity axis, while ignoring force and deformation in all other axes. To infer the joint parameters of coupled branches, we only measure the trajectories of the child branch to which the arm is attached, obviating the need for any visual tracking. We utilize the end-effector force readings exposed by the arm, which are typically inferred using forward kinematics, instead of the more accurate external sensor attachments at the end-effector itself, and consequently the readings are rather noisy. Additionally, the robot model could have position-sensitive bias in force readings, due to gravity compensation issues, sensor misalignment or improper arm calibration. However, we show that our framework is robust to these distortions. Both force and deformation measurements are independently retrieved from the arms using their respective Robot Operating System (ROS) APIs. Additionally, we neither use a perception mechanism to capture the tree characteristics nor do we attempt automated grasping.
 
Finally, we measure the branch attributes essential to create the digital replica, which are: the radius, length and density of the branch; and the angle at which the branch forks from its parent. While it is not necessary to simulate the entire tree, it is imperative that all upper level branches that participate in the collective dynamics of the target branch are accounted for. The density of wood material is assumed to be constant for a tree species (or the synthetic material), which can be derived indirectly from the mass and volumetric measurements of a single cut out fragment.

\subsection{Evaluation Metrics}
\label{evaluation-metrics}
To assess the estimation reliability, first we compare the inferred parameter distribution with the pre-specified, parameter values from the ground truth, using the sim-to-sim setup. In both sim-to-sim and real-to-sim experiments, we compare the unseen ground truth (GT) deformation trajectory to the predicted samples drawn from the estimated trajectory distribution $p(\tau_{sim} | \theta^{*})$ corresponding to the converged particle set $\theta^{*}$.  Additionally, in the latter case, we compute the 95\% confidence interval (CI) over the estimated trajectories and assess the fraction of true trajectory points that fall outside the interval. The distribution itself is constructed from the samples of trajectories corresponding to $\theta^{*}$, leveraging Kernel Density Estimation with Gaussian kernels and Scott's rule of thumb for bandwidth selection. In the sim-to-sim experiments, we use a hold-out  dataset for evaluation, while in real-to-sim, we use k-fold cross validation due to the smaller sample size. In either case, the evaluation trajectories are unseen during the parameter inference workflow. Samples of GT force-deformation measurements, along with all experiment details, such as the dataset volume and the hyper-parameters used are available in the supplementary material.

\section{Experiments and Results}

We assess the quality of our approach using the following experimental scenarios. 

\subsection{Baselines} 
\begin{figure}[htp]
    \centering
    \includegraphics[width=9.8cm]{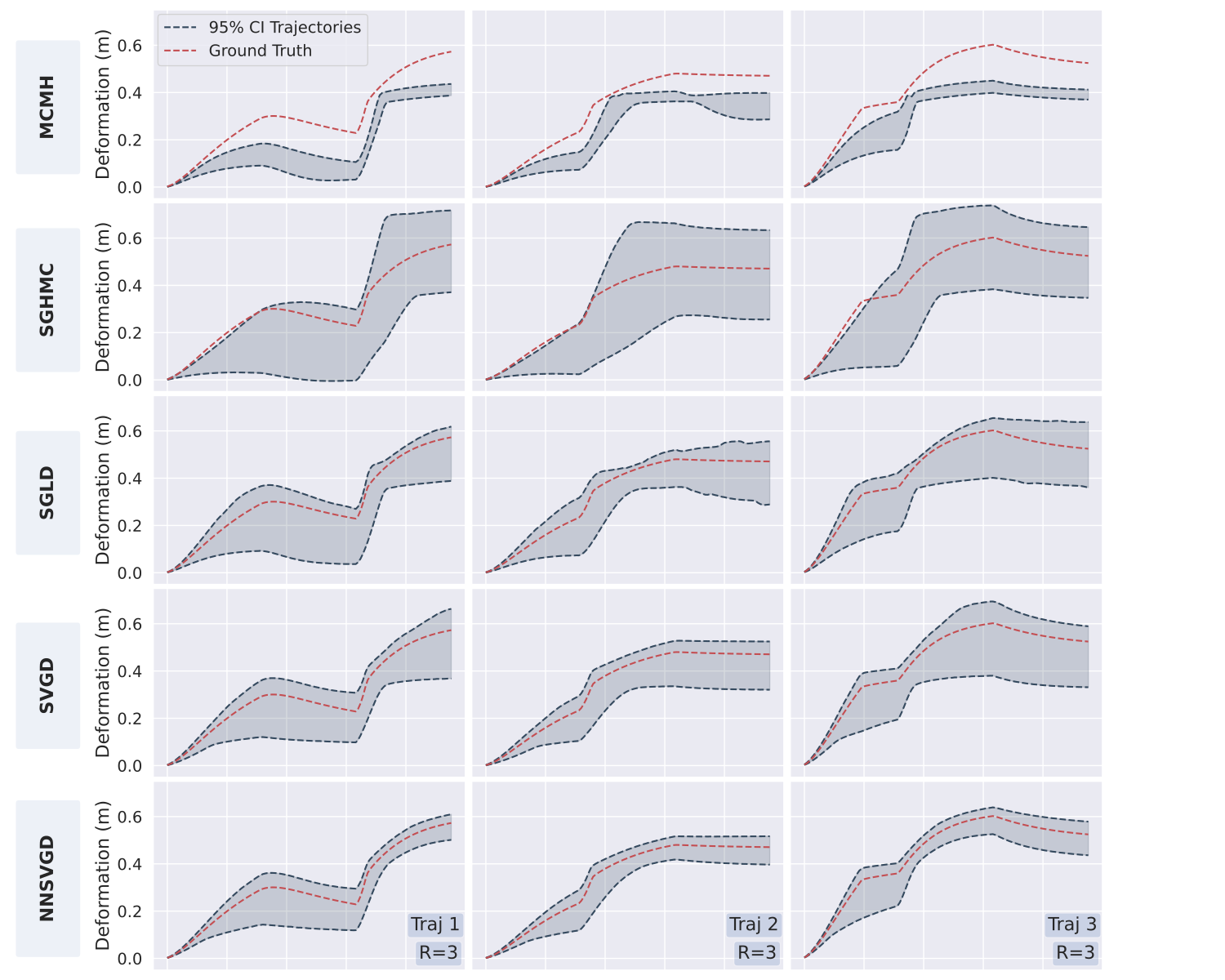}
    \caption{Comparison of NNSVGD (ours) against baseline inference algorithms. 95\% CI of the predicted distribution of test trajectories $p(\tau_{sim} | \theta^{*})$, computed with $R=3$, is shown in grey against the preset ground truth $\tau_{gt}$ in red}.
    \label{fig:comparison_metrics}
\end{figure}

In the first scenario, illustrated in Fig. \ref{fig:comparison_metrics}, we baseline our inference approach, SVGD with a neural network prior (NNSVGD), against four alternatives, which are: a) the classical Metropolis–Hastings (MCMH) algorithm b) Stochastic gradient Hamiltonian Monte Carlo (SGHMC) method c) Stochastic Gradient Langevin dynamics (SGLD) and a basic d)  Stein Variational Gradient Descent (SVGD). In each of the four baseline algorithms, we use the smooth box prior described in section \ref{smooth_box_prior} to set the search limits, and wherever a gradient is required, we use the finite difference based approximation. This scenario is executed using the sim-to-sim setup with a multi-branch simulated tree geometry ($R=3$), imitating a joint grandparent-parent-child motion. We keep all hyper-parameters, such as the number of iterations and the particle count, similar across the algorithms. In the gradient based methods, we use an Adam optimizer with a small learning rate to guide the descent.

The results in  Fig. \ref{fig:comparison_metrics} indicate all gradient based methods outperform the random-walk based MCMH. Furthermore, samples from the Stein methods, i.e., SVGD and NNSVGD, are better distributed around the ground truth $\tau_{gt}$ trajectory, while additionally displaying sharper convergence and lower prediction uncertainty, indicated by the CI width. For identical iteration counts, Stein methods outperform chain based SGHMC and SGLD chiefly because the particles are updated in parallel, a feature exploited by our parallel simulator.

\subsection{NNSVGD vs SVGD}

\begin{figure}[htp]
    \centering
    \includegraphics[width=8.7cm]{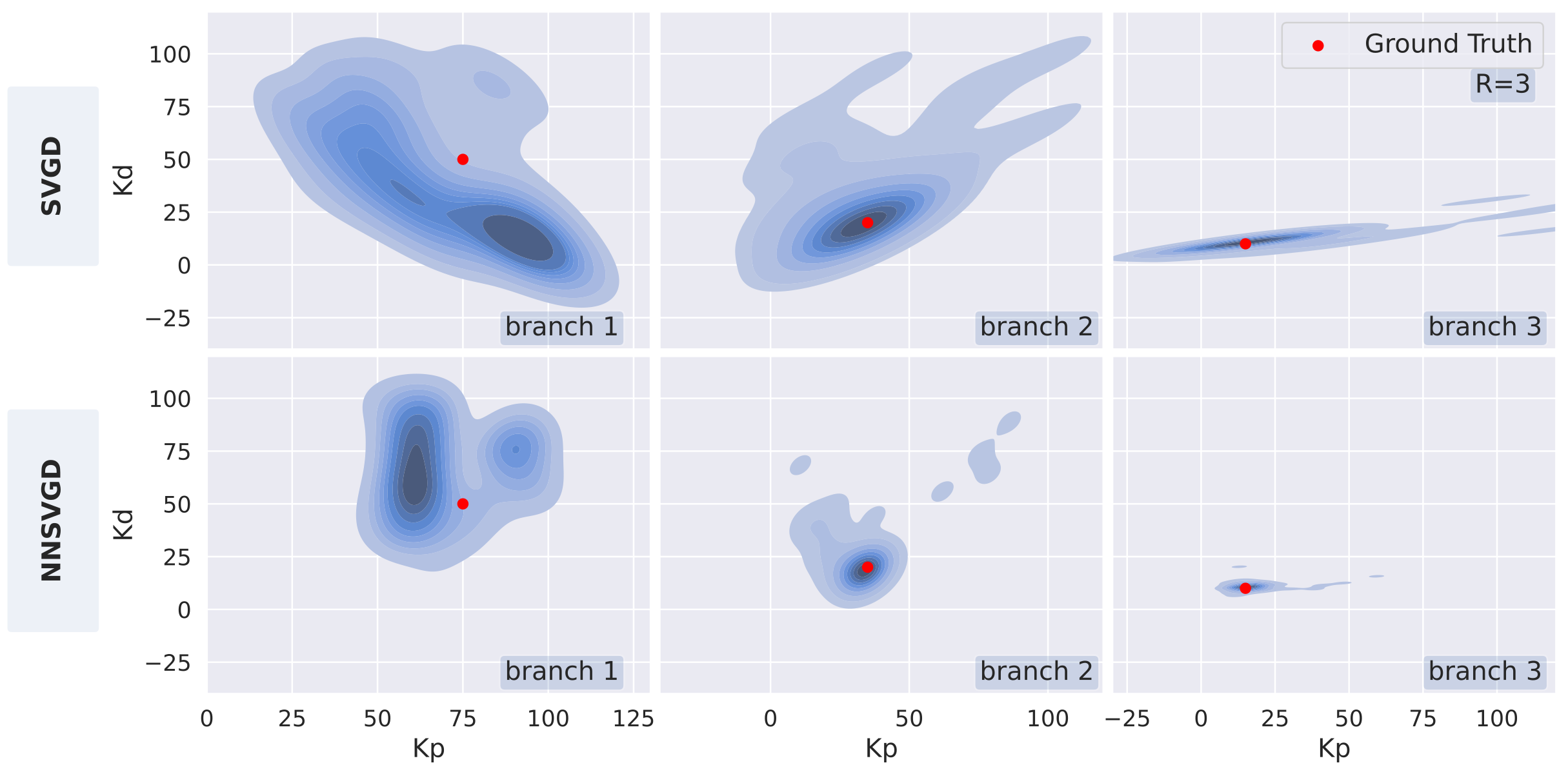}
    \caption{Joint posteriors with and without the neural network prior, computed with $R=3$. The red dots indicate the preset ground truth for stiffness $K_{p}$ and damping $K_{d}$ parameters.}
    \label{fig:nn_prior_benefit}
\end{figure}

With the same setup, the benefit of having a neural network prior, described in section \ref{nn_ineq_prior}, is examined both visually and quantitatively. First, the intuitive visual result in Fig. \ref{fig:nn_prior_benefit}, shows that the search space restriction from the inequality prior forces the spring parameters to have a sharper convergence, for all three branches. Consequently, the sample trajectories predicted by NNSVGD show lower CI widths in comparison to SVGD, from Fig. \ref{fig:comparison_metrics}. Furthermore, the multi-modal nature of the branch1 posterior indicates the presence of multiple solutions, a quintessential feature of inverse inference.

In the first quantitative test, we compute the negative log-likelihood (Nll) of the estimated posterior from 5 random training trajectories. Secondly, we compute the 95\% CI of the predicted trajectories averaged over 25 unseen test trajectories and assess the fraction of GT points that fall outside the interval, as detailed in section \ref{evaluation-metrics}.

\begin{table}[ht]
\centering
\scalebox{0.9}{
\begin{tabular}{c c c c} 
 \hline
   & 95\% CI width & \% GT outside CI & Nll \\ [0.2ex] 
 \hline\hline
 NNSVGD & 0.11 & $<$1\% & 2268.1 \\
 SVGD & 0.19 & $<$1\% & 3050.7 \\ [0.2ex] 
 \hline
\end{tabular}
}
\caption{NN prior benefits: Quantitative metrics.}
\label{table:nnsvgd_stat_metrics_table}
\end{table}

The results, in Table \ref{table:nnsvgd_stat_metrics_table} demonstrate that the addition of the neural network prior results in a lower Nll and a lower CI width for the same \%GT enclosure, suggesting its superiority.

\subsection{Robustness}
\begin{figure}[htp]
    \centering
    \includegraphics[width=9.0cm]
    {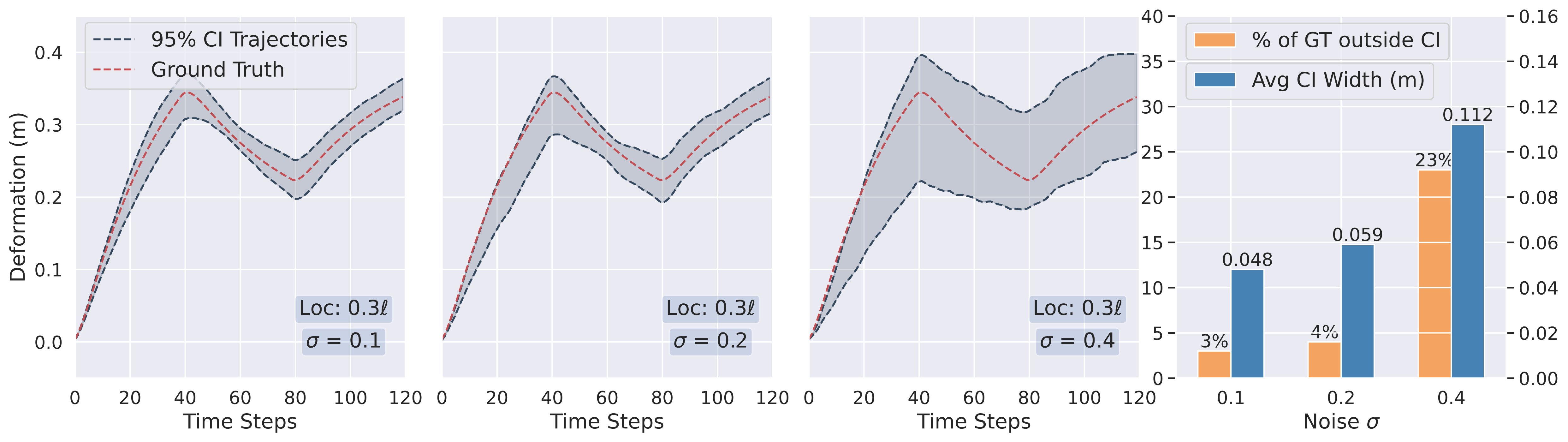}
    \caption{95\% CI of predicted distribution against GT for varying noise $\sigma$, measured relative to the force, keeping the grasp location fixed. The bar chart shows the CI width along with \%GT points outside the interval (lesser the better for both).}
    \label{fig:robustness_noise}
\end{figure}

In the second scenario, we perform two sets of experiments to measure the robustness of the model against erratic sensors and variations in branch grasp locations. In both sets, the metrics  are computed as an average over 25 unseen test force-deformation trajectories, 3 of which are shown along with the results. In the first set, we inject Gaussian noise of standard deviation $\sigma$ into the ground truth deformation and force trajectories, leveraging the sim-to-sim setup. The results in Fig. \ref{fig:robustness_noise}, show that for moderate levels of noise, the predicted distribution is well-aligned to the ground truth; furthermore, the 95\% CI attributes indicate that the model is able to capture the injected noise within the estimated uncertainty.

\begin{figure}[htp]
    \centering
    \includegraphics[width=9.0cm]
    {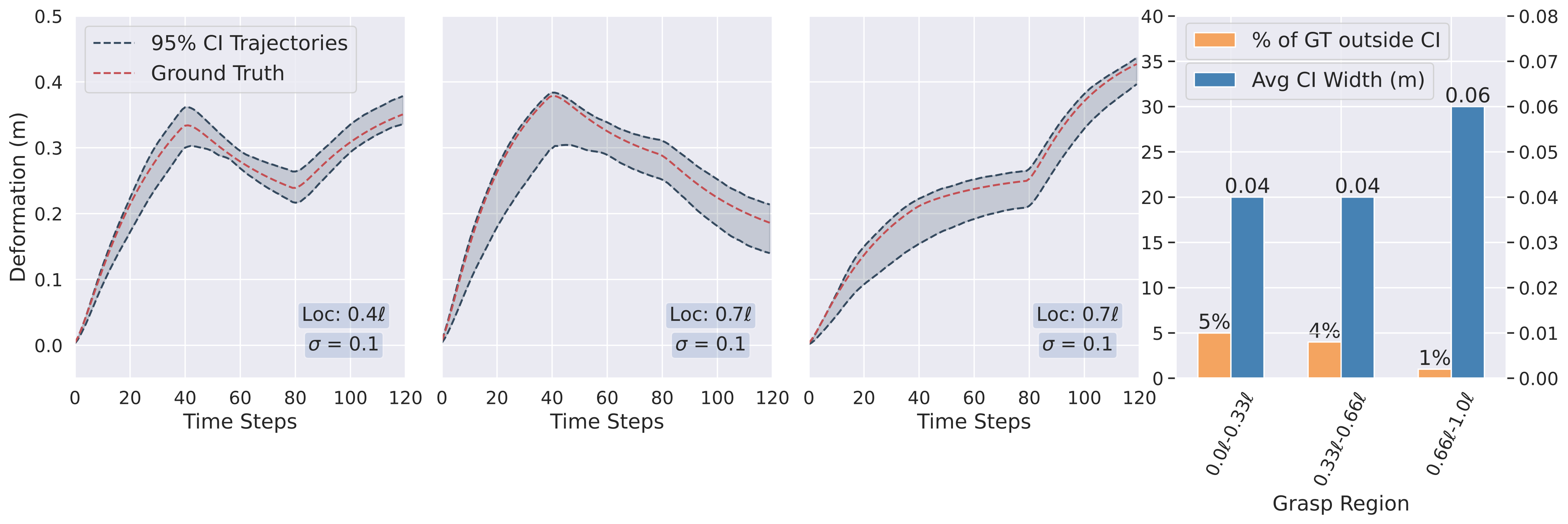}
    \caption{95\% CI of predicted distribution against GT for varying grasp locations, keeping noise $\sigma$ fixed. The grasp location is represented as a fraction of the branch length, from the parent joint; for e.g., Loc: 0.75$\ell$ is the mid point of the outer half.}
    \label{fig:robustness_location}
\end{figure}

Secondly, we vary the grasp locations, selected randomly along the branch, where the force is applied. This experiment mimics the situation where the arm needs to ascertain whether to grasp towards the outer edge of a branch to reduce the applied forces, or move towards the inner fork due to the arm reach constraints. The obtained results, illustrated in Fig. \ref{fig:robustness_location}, confirm that our model can successfully infer the deformation on one location using trajectories obtained by grasping another.





\subsection{Hardware Experiments}

\begin{figure}[htp]
  \centering
   {\includegraphics[width=8.9cm]{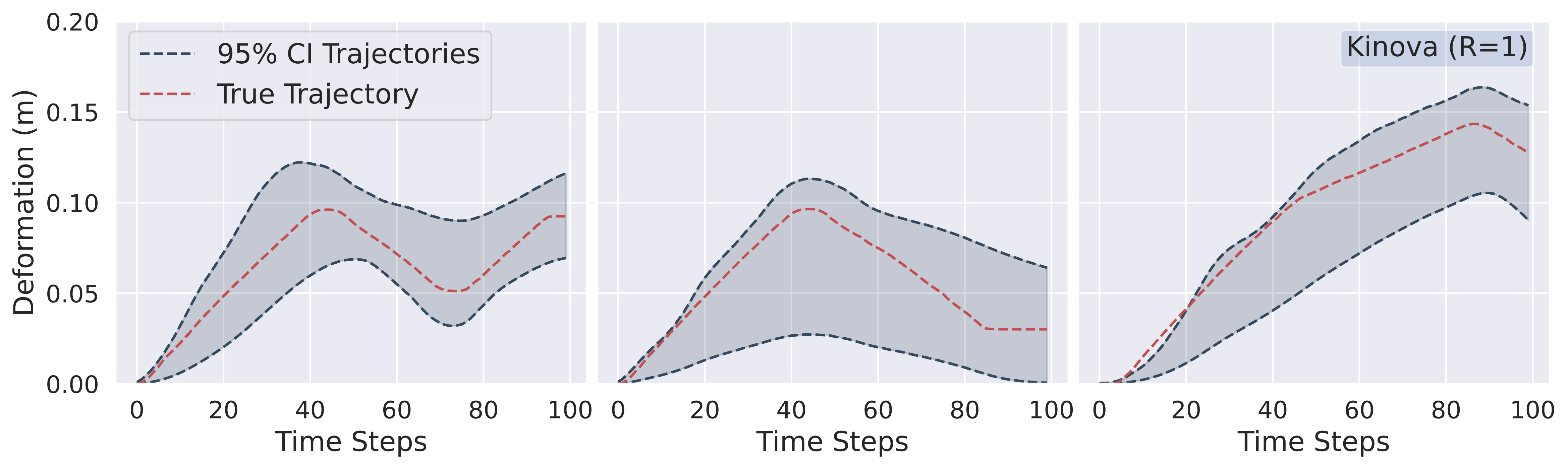}\label{fig: kinova_predicted_ci}}
   {\includegraphics[width=8.9cm]{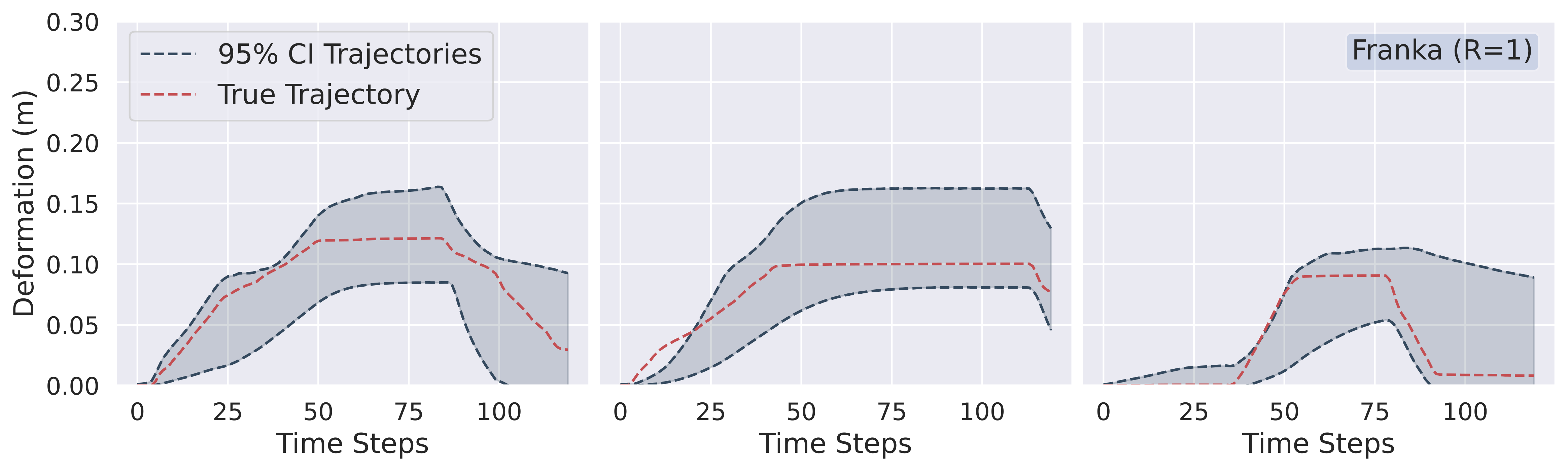}\label{fig:franka_predicted_ci}}
    {\includegraphics[width=8.9cm]{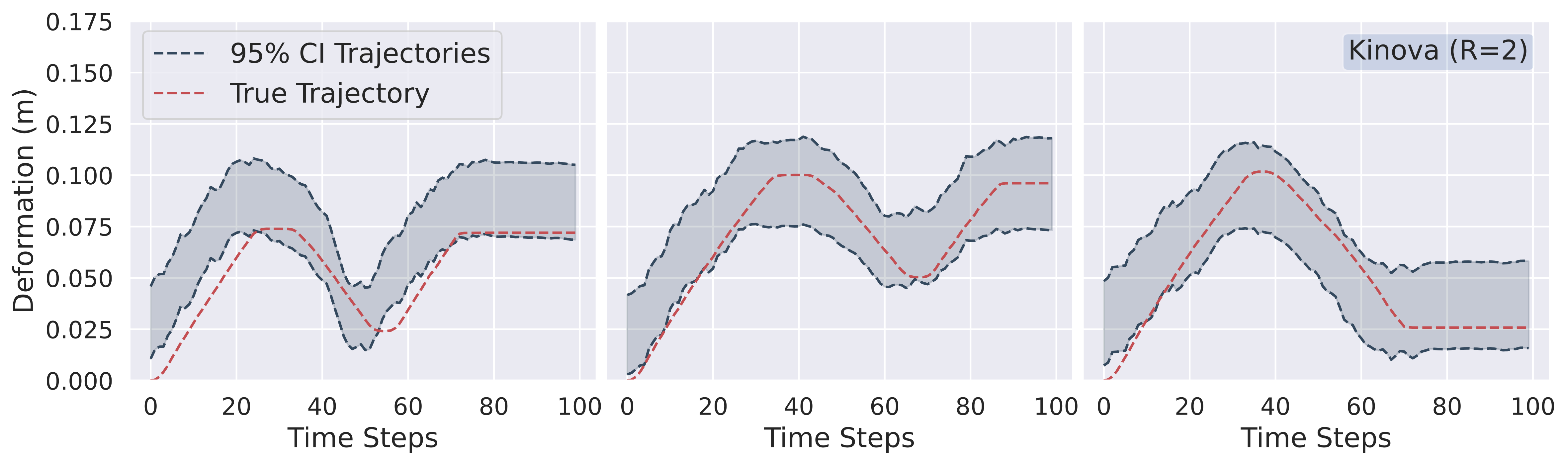}\label{fig:franka_predicted_ci}}
  \hfill
   \vspace*{-2mm}
  \caption{Real-to-sim results. Top: Kinova ($R=1$), Middle: Franka  ($R=1$), Bottom: Kinova   ($R=2$). Shaded regions represent the 95\% CI of the predicted distribution $p(\tau_{sim} | \theta^{*})$.}
  \label{fig:predicted_ci}
 \vspace*{-5mm}
\end{figure}

Here, we perform real-world experiments and capture multiple pairs of force-deformation trajectories using the two setups described in section \ref{real-to-sim}. The converged NNSVGD particles and the corresponding deformation trajectory samples are used to construct a trajectory distribution $p(\tau_{sim} | \theta^{*})$ and the 95\% CI for each time-step, shown in Fig. \ref{fig:predicted_ci}. The CI attributes are tabulated in Table. \ref{table:real_ci_metrics_table}.


\begin{table}[ht]
\centering
\scalebox{0.9}{
\begin{tabular}{ l c c c}
\hline
  & {Kinova ($R=1$)} & {Franka ($R=1$)}  & {Kinova ($R=2$)}  \\
\hline
\hline
Avg 95\% CI width  & 0.052 & 0.066 & 0.039 \\

Avg \%GT outside CI & 5.0\% & 7.0\% & 23.9\%  \\
\hline
\end{tabular}
}
\caption{Real-to-sim CI attributes }
\label{table:real_ci_metrics_table}
\end{table}

The results show that despite the heteroschedastic noise in the ground truth, our model is able to predict the deformation quite well. For instance, across the three farm tree trajectories, on average, 93.0\% of true points lie within the 95\% confidence interval and the average interval width is 6.6cm. We deem such metrics as acceptable for a majority of practical applications discussed earlier. Remarkably, these high quality real-world predictions are obtained by using as little as three training trajectories. However, from the real-world branch behaviour, we observed that our approach works the best for deformation in regions that are closer to the horizontal ($+45^{\degree}$ to  $-45^{\degree}$ from the ground axis). Given the force is applied along the gravity axis, this corresponds to a region where its normal component significantly outweighs the sheer. Therefore, the proposed method could be improved by using the branch angles captured through perception to apply normal forces to the branch instead, or alternatively, by representing a single real branch with multiple links to increase pliability, both of which we leave aside for future consideration.




\section{Conclusion}

In this work, we have demonstrated a method to learn the dynamic behaviour of tree branches by utilising the simulation-based parameter inference approach. In addition, we described a method for injecting biological assumptions into the Bayesian model via a neural network prior. Previous works on tree dynamics are either focused on perception or on the biological aspects of the tree; therefore, they are not designed for active branch manipulation, a key requirement for robotic harvesting and locomotion tasks. The comprehensive results indicate a robust model that captures the parameter uncertainty quite well even under noise and grasp location perturbations. Future work in this direction are as follows: First, the spring model can be alternatively built from a  perception acquired branch topology \cite{lowe2022tree}. Second, a generalised version of the model could be used to learn  branch manipulation policies. Third, severe discrepancies between model estimates and true deformation behaviour could be used as an indicator of poor tree health or branch rupture. Overall, we foresee works that entail combining our uncertainty aware model with visual tree reconstruction works, generalising it to different branches, and eventually generating robust control policies for manipulation.


\section*{Acknowledgment}
This work was (partially) funded by CSIRO's Reimagine Farming project part of CSIRO’s Reinvent Science program. The authors are grateful for technical and systems support from Tea Molnar, Brendan Tidd, Lauren Hanson and Peyman Moghadam on CSIRO's mobile manipulation hardware and software integration for field testing.


\bibliographystyle{IEEEtran}
\bibliography{IEEEabrv,main}

\end{document}